\documentclass[11pt,a4paper]{article}
\usepackage[hyperref]{acl2021}
\usepackage{times}
\usepackage{latexsym}

\usepackage{microtype}
\usepackage{graphicx}
\usepackage{amsmath}
\usepackage{booktabs}
\usepackage{multicol}
\usepackage{multirow}
\usepackage{ulem}
\usepackage{float}
\usepackage{url}
\usepackage{makecell}

\usepackage{color}
\definecolor{Orange}{RGB}{237,125,49}
\definecolor{Green}{RGB}{0,176,80}
\definecolor{Blue}{RGB}{0,112,192}

\newtheorem{definition}{Definition}

\aclfinalcopy % Uncomment this line for the final submission
\def\aclpaperid{2131} %  Enter the acl Paper ID here

\newcommand\modelname{TWAG}
\newcommand\modelnames{TWAG\ }
\newcommand\modelnameb{\textbf{TWAG}}
\newcommand\modelnamebs{\textbf{TWAG}\ }
\title{TWAG: A Topic-guided Wikipedia Abstract Generator}

\author{Fangwei Zhu$^{1,2}$, Shangqing Tu$^3$, Jiaxin Shi$^{1,2}$, Juanzi Li$^{1,2}$, Lei Hou$^{1,2}$\thanks{\quad Corresponding Author}\hspace{0.5em} and Tong Cui$^4$ \\
  $^1$Dept. of Computer Sci.\&Tech., BNRist, Tsinghua University, Beijing 100084, China \\
  $^2$KIRC, Institute for Artificial Intelligence, Tsinghua University \\
  $^3$School of Computer Science and Engineering, Beihang University \\
  $^4$Noah's Ark Lab, Huawei Inc. \\
  \texttt{\{zfw19@mails.,shijx16@mails,lijuanzi@,houlei@\}tsinghua.edu.cn} \\
  \texttt{tsq@buaa.edu.cn,cuitong5@huawei.com} \\
}

\date{\today}

\begin{document}
\maketitle
\begin{abstract}
Wikipedia abstract generation aims to distill a Wikipedia abstract from web sources and has met significant success by adopting multi-document summarization techniques.
However, previous works generally view the abstract as plain text, ignoring the fact that it is a description of a certain entity and can be decomposed into different topics.
In this paper, we propose a two-stage model \modelnames that guides the abstract generation with topical information.
First, we detect the topic of each input paragraph with a classifier trained on existing Wikipedia articles to divide input documents into different topics.
Then, we predict the topic distribution of each abstract sentence, and decode the sentence from topic-aware representations with a Pointer-Generator network.
We evaluate our model on the WikiCatSum dataset, and the results show that \modelnames outperforms various existing baselines and is capable of generating comprehensive abstracts.
Our code and dataset can be accessed at \url{https://github.com/THU-KEG/TWAG}
\end{abstract}

\section{Introduction}
\label{sec:intro}
Wikipedia, one of the most popular crowd-sourced online knowledge bases, has been widely used as the valuable resources in natural language processing tasks such as knowledge acquisition~\cite{lehmann2015dbpedia} and question answering~\cite{hewlett2016wikireading, rajpurkar2016squad} due to its high quality and wide coverage. 
Within a Wikipedia article, its abstract is the overview of the whole content, and thus becomes the most frequently used part in various tasks. However, the abstract is often contributed by experts, which is labor-intensive and prone to be incomplete.

In this paper, we aim to automatically generate Wikipedia abstracts based on the related documents collected from referred websites or search engines, which is essentially a multi-document summarization problem. This problem is studied in both extractive and abstractive manners. 

The extractive models attempt to select relevant textual units from input documents and combine them into a summary. 
Graph-based representations are widely exploited to capture the most salient textual units and enhance the quality of the final summary~\cite{erkan2004lexrank, mihalcea2004textrank, wan2008exploration}.
Recently, there also emerge neural extractive models~\cite{yasunaga2017graph, yin2019graph} utilizing the graph convolutional network~\cite{kipf2016semi} to better capture inter-document relations.
However, these models are not suitable for Wikipedia abstract generation. The reason is that the input documents collected from various sources are often noisy and lack intrinsic relations~\cite{sauper2009automatically}, which makes the relation graph hard to build.

The abstractive models aim to distill an informative and coherent summary via sentence-fusion and paraphrasing~\cite{filippova2008sentence, banerjee2015multi, bing2015abstractive}, but achieve little success due to the limited scale of datasets.
~\citet{liu2018generating} proposes an extractive-then-abstractive model and contributes WikiSum, a large-scale dataset for Wikipedia abstract generation, inspiring a branch of further studies~\cite{perez2019generating, liu2019hierarchical, li2020leveraging}.

\begin{figure}[t]
    \centering
    \includegraphics[width=\columnwidth]{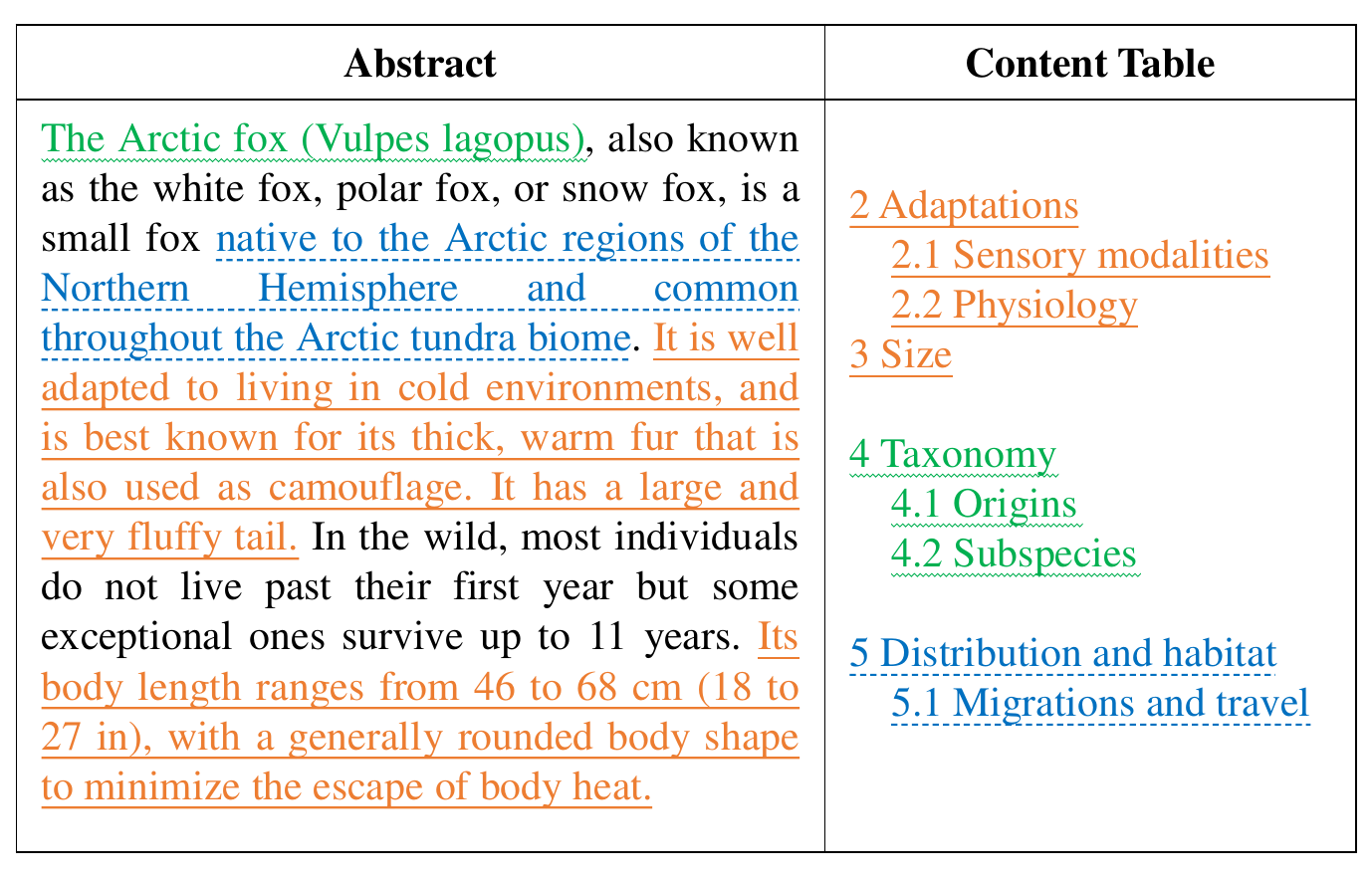}
    \caption{An example of Wikipedia article \textit{Arctic Fox}. The abstract contains three orthogonal topics about an animal: \uline{\textcolor{Orange}{Description}}, \uwave{\textcolor{Green}{Taxonomy}} and \dashuline{\textcolor{Blue}{Distribution}}. The right half is part of the article's content table, showing section labels related to different topics.}
    \label{fig:content_table}
\end{figure}

The above models generally view the abstract as plain text, ignoring the fact that Wikipedia abstracts describe certain entities, and the structure of Wikipedia articles could help generate comprehensive abstracts.
We observe that humans tend to describe entities in a certain domain from several topics when writing Wikipedia abstracts.
As illustrated in Figure \ref{fig:content_table}, the abstract of the \textit{Arctic Fox} contains its adaption, biology taxonomy and geographical distribution, which is consistent with the content table.
Therefore, given an entity in a specific domain, generating abstracts from corresponding topics would reduce redundancy and produce a more complete summary.

In this paper, we try to utilize the topical information of entities within its domain (Wikipedia categories) to improve the quality of the generated abstract. 
We propose a novel two-stage Topic-guided Wikipedia Abstract Generation model (\modelnameb).
\modelnames first divides input documents by paragraph and assigns a topic for each paragraph with a classifier-based topic detector.
Then, it generates the abstract in a sentence-wise manner, i.e., predicts the topic distribution of each abstract sentence to determine its topic-aware representation, and decodes the sentence with a Pointer-Generator network~\cite{see2017get}.
We evaluate \modelnames on the \textbf{WikiCatSum}~\cite{perez2019generating} dataset, a subset of the \textbf{WikiSum} containing three distinct domains. 
Experimental results show that it significantly improves the quality of abstract compared with several strong baselines. 

In conclusion, the contributions of our work are as follows:
\begin{itemize}
     \item We propose \modelname, a two-stage neural abstractive Wikipedia abstract generation model utilizing the topic information in Wikipedia, which is capable of generating comprehensive abstracts.
    \item We simulate the way humans recognize entities, using a classifier to divide input documents into topics, and then perform topic-aware abstract generation upon the predicted topic distribution of each abstract sentence.
    \item Our experiment results against 4 distinct baselines prove the effectiveness of TWAG.
\end{itemize}

\section{Related Work}
\label{sec:rw}
\subsection{Multi-document Summarization}
Multi-document summarization is a classic and challenging problem in natural language processing, which aims to distill an informative and coherent summary from a set of input documents.
Compared with single-document summarization, the input documents may contain redundant or even contradictory information~\cite{radev2000common}.

Early high-quality multi-document summarization datasets are annotated by humans, e.g., datasets for Document Understanding Conference (DUC) and Text Analysis Conference (TAC).
These datasets are too small to build neural models, and most of the early works take an extractive method, attempting to build graphs with inter-paragraph relations and choose the most salient textual units. The graph could be built with various information, e.g., TF-IDF similarity~\cite{erkan2004lexrank}, discourse relation~\cite{mihalcea2004textrank}, document-sentence two-layer relations~\cite{wan2008exploration}, multi-modal~\cite{wan2009graph} and query information~\cite{cai2012mutually}.

Recently, there emerge attempts to incorporate neural models, e.g., 
~\citet{yasunaga2017graph} builds a discourse graph and represents textual units upon the graph convolutional network (GCN)~\cite{kipf2016semi}, and ~\citet{yin2019graph} adopts the entity linking technique to capture global dependencies between sentences and ranks the sentences with a neural graph-based model.

In contrast, early abstractive models using sentence-fusion and paraphrasing~\cite{filippova2008sentence, banerjee2015multi, bing2015abstractive} achieve less success.
Inspired by the recent success of single-document abstractive models~\cite{see2017get, paulus2018deep, gehrmann2018bottom, huang2020knowledge}, some works~\cite{liu2018generating, zhang2018towards} try to transfer single-document models to multi-document settings to alleviate the limitations of small-scale datasets. 
Specifically, \citet{liu2018generating} defines Wikipedia generation problem and contributes the large-scale WikiSum dataset. 
\citet{fabbri2019multi} constructs a middle-scale dataset named MultiNews and proposes an extractive-then-abstractive model by appending a sequence-to-sequence model after the extractive step.
\citet{li2020leveraging} models inter-document relations with explicit graph representations, and incorporates pre-trained language models to better handle long input documents.

\subsection{Wikipedia-related Text Generation}
\label{sec: wiki_related_work}
\citet{sauper2009automatically} is the first work focusing on Wikipedia generation, which uses Integer Linear Programming (ILP) to select the useful sentences for Wikipedia abstracts.
\citet{banerjee2016wikiwrite} further evaluates the coherence of selected sentences to improve the linguistic quality.

\citet{liu2018generating} proposes a two-stage extractive-then-abstractive model, which first picks paragraphs according to TF-IDF weights from web sources, then generates the summary with a transformer model by viewing the input as a long flat sequence. 
Inspired by this work, \citet{perez2019generating} uses a convolutional encoder and a hierarchical decoder, and utilizes the Latent Dirichlet Allocation model (LDA) to render the decoder topic-aware. 
HierSumm~\cite{liu2019hierarchical} adopts a learning-based model for the extractive stage, and computes the attention between paragraphs to model the dependencies across multiple paragraphs.
However, these works view Wikipedia abstracts as plain text and do not explore the underlying topical information in Wikipedia articles. 

There are also works that focus on generating other aspects of Wikipedia text.
\citet{biadsy2008unsupervised} utilizes the key-value pairs in Wikipedia infoboxes to generate high-quality biographies.
\citet{hayashi2020wikiasp} investigates the structure of Wikipedia and builds an aspect-based summarization dataset by manually labeling aspects and identifying the aspect of input paragraphs with a fine-tuned RoBERTa model~\cite{liu2019roberta}. 
Our model also utilizes the structure of Wikipedia, but we generate the compact abstract rather than individual aspects, which requires the fusion of aspects and poses a greater challenge to understand the connection and difference among topics.

\begin{figure*}[htbp]
    \centering
    \includegraphics[width=\linewidth]{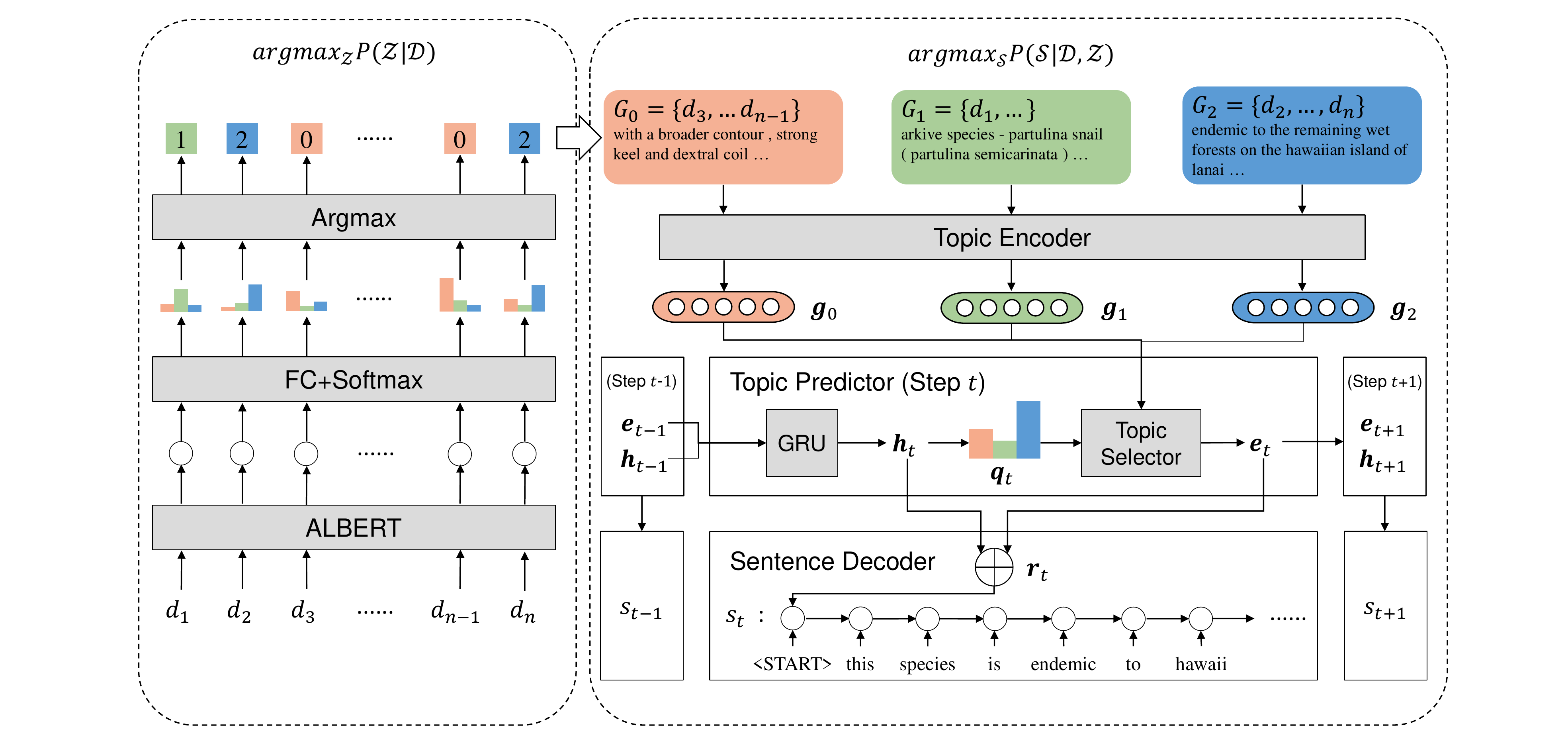}
    \caption{The \modelnames framework. We use an example domain with 3 topics for illustration. The left half is the topic detector which attempts to find a topic for each input paragraph, and the right half is the topic-aware abstract generator to generate the abstract by sentence based on input paragraphs and their predicted topics.}
    \label{fig:full_model}
\end{figure*}

\section{Problem Definition}
\label{sec:definition}

\begin{definition}
\textbf{Wikipedia abstract generation} accepts a set of paragraphs\footnote{The input documents can be represented by textual units with different granularity, and we choose paragraph as it normally expresses relatively complete and compact semantics.} $\mathcal{D} = \{d_1, d_2, \ldots, d_n\}$ of size $n$ as input, and outputs a Wikipedia abstract $\mathcal{S} = (s_1, s_2, \ldots, s_m)$ with $m$ sentences. The goal is to find an optimal abstract $\mathcal{S}^{*}$ that best concludes the input, i.e.,
\begin{equation}
\small
    \mathcal{S}^{*} = \mathop{\arg\max}_{\mathcal{S}} P(\mathcal{S}|\mathcal{D})
\end{equation}
\end{definition}

Previous works generally view $\mathcal{S}$ as plain text, ignoring the semantics in Wikipedia articles.
Before introducing our idea, let's review how Wikipedia organizes articles.

Wikipedia employs a hierarchical open category system to organize millions of articles, and we name the top-level category as domain.
As for a Wikipedia article, we concern three parts, i.e., the abstract, the content table, and textual contents. Note that the content table is composed of several section labels $\{l\}$, pairing with corresponding textual contents $\{p\}$.
As illustrated in Figure \ref{fig:content_table}, the content table indicates different aspects (we call them topics) of the article, and the abstract semantically corresponds to these topics, telling us that topics could benefit the abstract generation.

However, general domains like \textit{Person} or \textit{Animal} consist millions of articles with diverse content tables, making it not feasible to simply treat section labels as topics. Considering that articles in specific domains often share several salient topics, we manually merge similar section labels to convert the sections titles to a set of topics. Formally, the topic set is denoted as $\mathcal{T} = \{T_1, T_2, ..., T_{n_t}\}$ of size $n_t$, where each topic $T_i = \{l_{i}^{1}, l_{i}^{2}, \ldots, l_{i}^{m}\}$.

Now, our task can be expressed with a topical objective, i.e., 
\begin{definition}
Given the input paragraphs $\mathcal{D}$, we introduce the latent topics $\mathcal{Z} = \{z_1, z_2, \ldots, z_n\}$, where $z_i \in \mathcal{T}$ is the topic of $i$-th input paragraph $d_i$, and our objective of Wikipedia abstract generation is re-written as
\begin{equation}
\small
    \mathcal{S}^{*} = \mathop{\arg\max}_{\mathcal{Z}} P(\mathcal{Z}|\mathcal{D}) \mathop{\arg\max}_{\mathcal{S}}P(\mathcal{S}|\mathcal{D}, \mathcal{Z}).
\end{equation}
\end{definition}
Therefore, the abstract generation could be completed with two sub-tasks, i.e., topic detection to optimize $\mathop{\arg\max}_{\mathcal{Z}} P(\mathcal{Z}|\mathcal{D})$ and topic-aware abstract generation to optimize $\mathop{\arg\max}_{\mathcal{S}}P(\mathcal{S}|\mathcal{D}, \mathcal{Z})$.

\section{The Proposed Method}
\label{sec:method}
As shown in Figure \ref{fig:full_model}, our proposed \modelnames adopts a two-stage structure. First, we train a topic detector based on existing Wikipedia articles to predict the topic of input paragraphs. Second, we group the input paragraphs by detected topics to encode them separately, and generate the abstract in a sentence-wise manner. In each step, we predict the topic distribution of the current sentence, fuse it with the global hidden state to get the topic-aware representation, and generate the sentence with a copy-based decoder. Next, we will detail each module.

\subsection{Topic Detection}
\label{sec:topic_detector}
The topic detector aims to annotate input paragraphs with their optimal corresponding topics.
To formalize, given the input paragraphs $\mathcal{D}$, $\text{Det}$ returns its corresponding topics $\mathcal{Z} = \{z_1, z_2, \ldots, z_n\}$, i.e.,
\begin{equation}
    \mathcal{Z} = \text{Det}(\mathcal{D})
\end{equation}

We view topic detection as a classification problem. 
For each paragraph $d \in \mathcal{D}$, we encode it with ALBERT\cite{lan2019albert} and then predict its topic $z$ with a fully-connected layer, i.e.,
\begin{align}
    \mathbf{d} &= \text{ALBERT}(d) \\
    z &= \arg\max(\text{linear}(\mathbf{d}))
\end{align}
where $\mathbf{d}$ is the vector representation of $d$, and we fine-tuned the ALBERT model on a pretrained version.

\subsection{Topic-aware Abstract Generation}
\label{sec:sentence_generator}
Topic-aware abstract generator utilizes the input paragraphs $\mathcal{D}$ and the detected topics $\mathcal{Z}$ to generate the abstract. 
Specifically, it contains three modules: a topic encoder to encode the input paragraphs into topical representations, a topic predictor to predict the topic distribution of abstract sentences and generate the topic-aware sentence representation, and a sentence decoder to generate abstract sentences based on the topic-aware representations.

\subsubsection{Topic Encoder}
Given the input paragraphs $\mathcal{D}$ and the detected topics $\mathcal{Z}$, we concatenate all paragraphs belonging to the same topic $T_k$ to form a topic-specific text group (TTG) $\mathcal{G}_k$, which contains salient information about a certain topic of an entity:
\begin{equation}
    \mathcal{G}_k = \text{concat}(\{d_i|z_i = T_k\}).
\end{equation}

To further capture hidden semantics, we use a bidirectional GRU to encode the TTGs:
\begin{equation}
    \mathbf{g}_k, \mathbf{U}_k = \text{BiGRU}(\mathcal{G}_k).
\end{equation}
$\mathbf{g}_k$ is the final hidden state of the $\mathcal{G}_k$, and $\mathbf{U}_k = (\mathbf{u}_1, \mathbf{u}_2,\ldots,\mathbf{u}_{n_{G_{k}}})$ represents the hidden state of each token in $\mathcal{G}_k$, where $n_{G_{k}}$ denotes the number of tokens in $\mathcal{G}_k$.

\subsubsection{Topic Predictor}
After encoding the topics into hidden states, \modelnames tackles the decoding process in a sentence-wise manner:
\begin{equation}
\small
    \mathop{\arg\max}_{\mathcal{S}} P(\mathcal{S}|\mathcal{D}, \mathcal{Z}) = \prod_{i=1}^{m}\mathop{\arg\max}_{s_i} P(s_i|\mathcal{D}, \mathcal{Z}, s_{<i})
\end{equation}

To generate the abstract $\mathcal{S}$, we first predict the topic distribution of every sentence $s_i$ with a GRU decoder. 
At each time step $t$, the topic predictor produces a global hidden state $\mathbf{h}_t$, and then estimates the probability distribution $\mathbf{q}_t$ over topics. 
\begin{align}
    \mathbf{h}_t &= \text{GRU}(\mathbf{h}_{t-1}, \mathbf{e}_{t-1}) \\
    \mathbf{q}_t &= \text{softmax}(\text{linear}({\mathbf{h}}_t))
\end{align}
where $\mathbf{e}_{t-1}$ denotes the topical information in the last step. $\mathbf{e}_0$ is initialized as an all-zero vector, and $\mathbf{e}_t$ could be derived from $\mathbf{q}_t$ in two ways.

The first way named \textbf{hard topic}, is to directly select the topic with the highest probability, and take its corresponding representation, i.e.,
\begin{equation}
    \mathbf{e}_{t}^{hard} = \mathbf{g}_{\mathop{\arg\max}_{i}(q_i)}.
\end{equation}
The second way named \textbf{soft topic}, is to view every sentence as a mixture of different topics, and take the weighted sum over topic representations, i.e.,
\begin{equation}
    \mathbf{e}_{t}^{soft} = \mathbf{q}_{t} \cdot \mathbf{G} 
\end{equation}
where $\mathbf{G} = (\mathbf{g}_{1}, \mathbf{g}_{2}, \ldots, \mathbf{g}_{n_t})$ is the matrix of topic representations.
With the observation that Wikipedia abstract sentences normally contain mixed topics, we choose the soft topic mechanism for our model (see Section \ref{sec:ablation_study} for details).

Finally, we compute the topic-aware hidden state $\mathbf{r}_t$ by adding up $\mathbf{h}_t$ and $\mathbf{e}_{t}$, which serves as the initial hidden state of sentence decoder:
\begin{equation}
    \mathbf{r}_{t} = \mathbf{h}_{t} + \mathbf{e}_{t}
\end{equation}

Additionally, a stop confirmation is executed at each time step:
\begin{equation}
    p_{stop} = \sigma(\text{linear}(\mathbf{h}_t))
\end{equation}
where $\sigma$ represents the sigmoid function.
If $p_{stop} > 0.5$, \modelnames will terminate the decoding process and no more abstract sentences will be generated.

\subsubsection{Sentence Decoder}
Our sentence decoder adopts the Pointer-Generator network~\cite{see2017get}, which picks tokens both from input paragraphs and vocabulary.

To copy a token from the input paragraphs, the decoder requires the token-wise hidden states $\mathbf{U} = (\mathbf{u}_1, \mathbf{u}_2, \ldots, \mathbf{u}_{n_u})$ of all $n_u$ input tokens, which is obtained by concatenating the token-wise hidden states of all TTGs, i.e.,
\begin{equation}
    \mathbf{U} = [\mathbf{U}_1, \mathbf{U}_2, \ldots, \mathbf{U}_{n_u}]
\end{equation}
For the $k$-th token, the decoder computes an attention distribution $\mathbf{a}_k$ over tokens in the input paragraphs, where each element $\mathbf{a}_{k}^{i}$ could be viewed as the probability of the $i$-th token being selected,
\begin{equation}
    \mathbf{a}_{k}^{i} = \text{softmax}(\tanh(\mathbf{W}_u \mathbf{u}_i+\mathbf{W}_s \mathbf{s}_k+\mathbf{b}_{a}))
\end{equation}
where $\mathbf{s}_k$ denotes the decoder hidden state with $\mathbf{s}_0 = \mathbf{r}_t$ to incorporate the topic-aware representation, and $\mathbf{W}_u, \mathbf{W}_s, \mathbf{b}_{a}$ are trainable parameters.

To generate a token from the vocabulary, we first use the attention mechanism to calculate the weighted sum of encoder hidden states, known as the context vector,
\begin{equation}
    \mathbf{c}^{*}_{k} = \sum_{i} \mathbf{a}_{k}^{i}\mathbf{u}_{i}.
\end{equation}
which is further fed into a two-layer network to obtain the probability distribution over vocabulary,
\begin{equation}
    P_{voc} = \text{softmax}(\text{linear}(\text{linear}([\mathbf{s}_k, \mathbf{c}^{*}_{k}]))).
\end{equation}

To switch between these two mechanisms, $p_{gen}$ is computed from context vector $\mathbf{c}^{*}_{k}$, decoder hidden state $\mathbf{s}_k$ and decoder input $\mathbf{x}_k$:
\begin{equation}
    p_{gen} = \sigma(\mathbf{W}_{c}^{T} \mathbf{c}^{*}_{k} + \mathbf{W}_{s}^{T} \mathbf{s}_k + \mathbf{W}_{x}^{T}\mathbf{x}_k + \mathbf{b}_{p})
\end{equation}
where $\sigma$ represents the sigmoid function and $\mathbf{W}_{c}^{T}, \mathbf{W}_{s}^{T}, \mathbf{W}_{x}^{T}$ and $\mathbf{b}_{p}$ are trainable parameters.
The final probability distribution of words is\footnote{$ww_{i}$ means the token corresponding to $\mathbf{u}_i$.}
\begin{equation}
\small
    P(w) = p_{gen}P_{voc}(w) + (1-p_{gen})\sum_{i:|ww_{i}=w}\mathbf{a}_{k}^{i}
\end{equation}

\subsection{Training}
\label{ssec:training}
The modules for topic detection and abstract generation are trained separately.

\subsubsection{Topic Detector Training}
\label{sssec:tdt}
Since there are no public benchmarks for assigning input paragraphs with Wikipedia topics, we construct the dataset with existing Wikipedia articles.
In each domain, we collect all the label-content pairs $\{(l, p)\}$ (defined in Section \ref{sec:definition}), and split the content into paragraphs $p = (d_1, d_2, \ldots, d_{n_p})$ to form a set of label-paragraph pairs $\{(l, d)\}$.
Afterwards, we choose all pairs $(l, d)$ whose section label $l$ belongs to a particular topic $T \in \mathcal{T}$ to complete the dataset construction, i.e., the topic-paragraph set $\{(T, d)\}$. Besides, a \textit{NOISE} topic is set up in each domain, which refers to meaningless text like scripts and advertisements, and the corresponding paragraphs are obtained by utilizing regular expressions to match obvious noisy texts. The details are reported in Appendix A.

Note that the dataset for abstract generation is collected from non-Wikipedia websites (refer to Section \ref{sec:exp} for details). These two datasets are independent of each other, which prevents potential data leakage.

In the training step, we use the negative log-likelihood loss to optimize the topic detector.
\begin{table*}[htbp]
    \centering
    \begin{tabular}{c|c c c c|c c c c}
         \toprule
         Domain & \#Examples & R1-r & R2-r & RL-r & \#Topics & Train & Valid & Test \\
         \midrule
         Company & 62,545 & .551 & .217 & .438 & 4 & 35,506 & 1,999 & 2,212 \\  
         Film    & 59,973 & .559 & .243 & .456 & 5 & 187,221 & 10,801 & 10,085\\
         Animal  & 60,816 & .541 & .208 & .455 & 4 & 51,009 & 2,897 & 2,876\\
         \bottomrule
    \end{tabular}
    \caption{Details about used datasets. The left half shows parameters about the WikiCatSum dataset: number of examples and ROUGE 1, 2, L recalls. The right half shows parameters about the dataset for topic detector: number of topics and number of topic-paragraph pairs in each split.}
    \label{tab:dataset}
\end{table*}

\subsubsection{Abstract Generator Training}
\label{sssec:agt}
The loss of topic-aware abstract generation step consists of two parts: the first part is the average loss of sentence decoder for each abstract sentence $\mathcal{L}_{sent}$, and the second part is the cross-entropy loss of stop confirmation $\mathcal{L}_{stop}$.

Following~\cite{see2017get}, we compute the loss of an abstract sentence by averaging the negative log likelihood of every target word in that sentence, and achieve $\mathcal{L}_{sent}$ via averaging over all $m$ sentences,
\begin{equation}
    \mathcal{L}_{sent} = \frac{1}{m}\sum^{m}_{t=1}\left(\frac{1}{n_{s_t}}\sum^{n_{s_t}}_{i=1} -\log P(w_i)\right)
\end{equation}
where $n_{s_t}$ is the length of the $t$-th sentence of the abstract. As for $\mathcal{L}_{stop}$, we adopt the cross-entropy loss, i.e.,
\begin{equation}
\small
    \mathcal{L}_{stop} = -y_{s}\log(p_{stop}) - (1-y_{s})\log(1-p_{stop})
\end{equation}
where $y_{s} = 1$ when $t > m$ and $y_{s} = 0$ otherwise.

\section{Experiments}
\label{sec:exp}

\subsection{Experimental Settings}

\paragraph{Dataset.} To evaluate the overall performance of our model, we use the \textbf{WikiCatSum} dataset proposed by~\cite{perez2019generating}, which contains three distinct domains (\textit{Company}, \textit{Film} and \textit{Animal}) in Wikipedia. 
Each domain is split into train (90\%), validation (5\%) and test (5\%) set. 

We build the dataset for training and evaluating the topic detector from the 2019-07-01 English Wikipedia full dump.
For each record in the WikiCatSum dataset, we find the article with the same title in Wikipedia dump, and pick all section label-content pairs $\{(l, p)\}$ in that article.
We remove all hyperlinks and graphics in contents, split the contents into paragraphs with the \textit{spaCy} library, and follow the steps in Section \ref{sssec:tdt} to complete dataset construction. Finally, we conduct an 8:1:1 split for train, validation and test.

Table \ref{tab:dataset} presents the detailed parameters of used datasets.

\paragraph{Evaluation Metrics.} We evaluate the performance of our model with ROUGE scores~\cite{lin2004rouge}, which is a common metric in comparing generated and standard summaries. 
Considering that we do not constrain the length of generated abstracts, we choose ROUGE F1 score that combines precision and recall to eliminate the tendency of favoring long or short results.

\paragraph{Implementation Details.} We use the open-source \textit{PyTorch} and \textit{transformers} library to implement our model.
All models are trained on NVIDIA GeForce RTX 2080.

In topic detection, we choose the top 20 frequent section labels in each domain and manually group them into different topics (refer to the Appendix A for details). For training, we use the pretrained \textit{albert-base-v2} model in the \textit{transformers} library, keep its default parameters and train the module for 4 epochs with a learning rate of 3e-5. 

For abstract generation, we use a single-layer BiGRU network to encode the TTGs into hidden states of 512 dimensions. 
The first 400 tokens of input paragraphs are retained and transformed into GloVe~\cite{pennington2014glove} embedding of 300 dimensions. 
The vocabulary size is 50000 and out-of-vocabulary tokens are represented with the average embedding of its adjacent 10 tokens. 
This module is trained for 10 epochs, the learning rate is 1e-4 for the first epoch and 1e-5 for the rest.

Before evaluation, we remove sentences that have an overlap of over 50\% with other sentences to reduce redundancy.

\begin{table*}[htbp]
    \centering
    \begin{tabular}{c|c c c|c c c|c c c}
        \toprule
        \multirow{2}{*}{Model} & \multicolumn{3}{c|}{Company} & \multicolumn{3}{c|}{Film} & \multicolumn{3}{c}{Animal} \\
        & R1 & R2 & RL & R1 & R2 & RL & R1 & R2 & RL \\
        \midrule 
        TF-S2S & .197 & .023 & .125 & .198 & .065 & .172 & .252 & .099 & .210 \\
        CV-S2D+T & .275 & .106 & .214 & .380 & \textbf{.212} & .323 & .427 & \textbf{.279} & .379 \\
        HierSumm & .133 & .028 & .107 & .246 & .126 & .185 & .165 & .069 & .134 \\
        BART & .310 & .116 & .244 & .375 & .199 & .325 & .376 & .226 & .335 \\
        TWAG (ours) & \textbf{.341} & \textbf{.119} & \textbf{.316} & \textbf{.408} & \textbf{.212} & \textbf{.343} & \textbf{.431} & .244 & \textbf{.409} \\
        \bottomrule
    \end{tabular}
    \caption{ROUGE F1 scores of different models.}
    \label{tab:rouge_results}
\end{table*}

\paragraph{Baselines.} We compare our proposed \modelnamebs with the following strong baselines:
\begin{itemize}
    \item \textbf{TF-S2S} \cite{liu2018generating} uses a Transformer decoder and compresses key-value pairs in self-attention with a convolutional layer.
    \item \textbf{CV-S2D+T} \cite{perez2019generating} uses a convolutional encoder and a two-layer hierarchical decoder, and introduces LDA to model topical information.
    \item \textbf{HierSumm} \cite{liu2019hierarchical} utilizes the attention mechanism to model inter-paragraph relations and then enhances the document representation with graphs.
    \item \textbf{BART} \cite{lewis2020bart} is a pretrained sequence-to-sequence model that achieved success on various sequence prediction tasks.
\end{itemize}

We fine-tune the pretrained BART-base model on our dataset and set beam size to 5 for all models using beam search at test time. 
The parameters we use for training and evaluation are identical to these in corresponding papers.

\subsection{Results and Analysis}

Table \ref{tab:rouge_results} shows the ROUGE F1 scores of different models. 
In all three domains, \modelnames outperforms other baselines.
Our model surpasses other models on ROUGE-1 score by a margin of about 10\%, while still retaining advantage on ROUGE-2 and ROUGE-L scores.
In domain \textit{Company}, our model boosts the ROUGE-L F1 score by about 30\%, considering that ROUGE-L score is computed upon the longest common sequence, the highest ROUGE-L score indicates that abstracts generated by \modelnames have the highest holistic quality.

While CVS2D+T and BART retain reasonable scores, TF-S2S and HierSumm do not reach the scores they claim in their papers.
Notice that the WikiCatSum dataset is a subset of WikiSum, which is used as the training dataset of these two models, we infer that TF-S2S and HierSumm require more training data to converge, and suffer from underfitting due to the dataset scale.
This phenomenon also proves that \modelnames is data-efficient.

\begin{table*}[htbp]
  \small
  \centering
    \begin{tabular}{|p{0.9\linewidth}|}
    \hline
    {\textbf{\textit{Gold Abstract:}} manjina there ( kannada : $\langle$unk$\rangle$ $\langle$unk$\rangle$ ) is a 1980 indian kannada film, directed by bangalore nagesh and produced by k. r. narayana murthy. the film stars srinath, manjula, thoogudeepa srinivas and dinesh in lead roles. the film had musical score by upendra kumar.} \\
    \hline
    {\textbf{\textit{TF-S2S:}} chalo ishq larain is a pakistani film directed by sajjad gul and was released across pakistan in april 2002 starring meera , ali haider and zara sheikh.} \\
    \hline
    {\textbf{\textit{CV-S2D+T:}} $\langle$unk$\rangle$ $\langle$unk$\rangle$ ( kannada : $\langle$unk$\rangle$ $\langle$unk$\rangle$ ) is a 1967 indian kannada film, directed by b. mallesh and produced by b. s. ranga. the film stars rajkumar, udaykumar, narasimharaju and k. s. ashwath in lead roles. the film had musical score by chellapilla satyam . the film was a remake of hindi film $\langle$unk$\rangle$.} \\
    \hline
    {\textbf{\textit{HierSumm:}} ( kannada : $\langle$unk$\rangle$ $\langle$unk$\rangle$ ) is a 1980 indian kannada film , directed by bangalore nagesh . the film stars srinath , manjulla , thoogudeepa shreenivaas . the film stars srinath , manjula , manjula , thoogudeepa shreenivaas in lead roles . the film had musical score by upendra kumar . the film is a remake of telugu movie aakashagiri rao . the movie was remade in telugu as manjina in 1986 . the movie was remade in telugu as manjina there \ldots} \\
    \hline
    {\textbf{\textit{BART:}} manjina there is a 1980 kannada family drama film directed by bangalore nagesh starring srinath and manjula in the lead roles. it was released on 14 january 1980.} \\
    \hline
    {\textbf{\textit{TWAG:}} manjina there is a 1980 kannada drama film directed by bangalore nagesh. the film stars srinath, vajramuni, manjula and thoogudeepa srinivas in lead roles. the film had musical score by upendra kumar and the film opened to positive reviews in 1980. the film was a remake of tamil film $\langle$unk$\rangle$.} \\
    \hline
    \end{tabular}
  \caption{Comparison between Wikipedia abstracts generated by different models about the film \textit{Majina There}. Non-English characters have been replaced with $\langle$unk$\rangle$ for readability.\label{tab:qualitative}}
\end{table*}

\subsection{Ablation Study}
\label{sec:ablation_study}

\paragraph{Learning Rate of Topic Detector.} We tried two learning rates when training the topic detector module.
A learning rate of 1e-7 would result in a precision of $0.922$ in evaluation, while a learning rate of 3e-5 would result in a precision of $0.778$.
However, choosing the former learning rate causes a drop of about 10\% in all ROUGE scores, which is the reason why we use the latter one in our full model.

We infer that human authors occasionally make mistakes, assigning paragraphs into section labels that belong to other topics.
A topic detector with low learning rate overfits these mistakes, harming the overall performance of our model.

\paragraph{Soft or Hard Topic.} To further investigate the effectiveness of \modelname's soft topic mechanism, we compare the results of soft and hard topic and report them in Table \ref{tab:topic_comparison}, from which we can see that hard topic does quite poorly in this task.
\begin{table}[H]
    \centering
    \scalebox{0.8}{
    \begin{tabular}{c|c c c|c c c}
        \toprule
        \multirow{2}{*}{Topic Detector} & \multicolumn{3}{c|}{Hard Topic} & \multicolumn{3}{c}{Soft Topic} \\ 
        & R1 & R2 & RL & R1 & R2 & RL \\
        \midrule 
        Company & .266 & .074 & .245 & \textbf{.341} & \textbf{.119} & \textbf{.316} \\
        Film & .355 & .159 & .333 & \textbf{.408} & \textbf{.212} & \textbf{.343} \\
        Animal & .407 & .223 & .387 & \textbf{.431} & \textbf{.244} & \textbf{.409} \\
        \bottomrule
    \end{tabular}
    }
    \caption{ROUGE F1 scores of different topic selectors.}
    \label{tab:topic_comparison}
\end{table}
A possible reason is that some sentences in the standard abstract express more than one topic. 
Assigning one topic to each sentence will result in semantic loss and thus harm the quality of generated abstract, while the soft topic could better simulate the human writing style.

\begin{figure}[htbp]
    \centering
    \includegraphics[width=\linewidth]{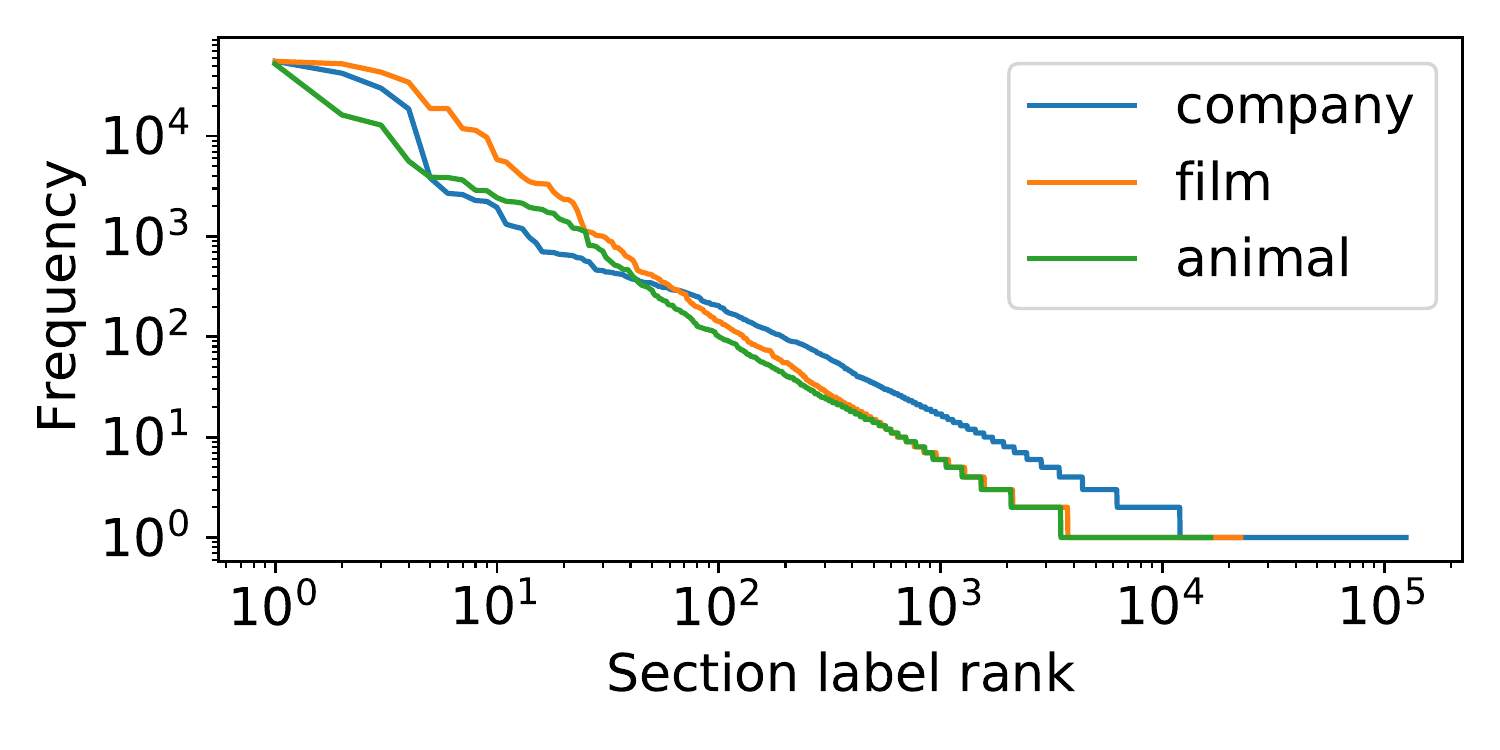}
    \caption{The frequency of section labels in three domains. When ignoring section labels with extra high or low frequency, remaining section labels' frequency and rank generally form a straight line in log scale, which matches the Zipf's law for long-tail distributions.}
    \label{fig:title_freq}
\end{figure}

\paragraph{Number of Section Labels.} The number of section labels $n_t$ plays a key role in our model: a small $n_t$ would not be informative enough to build topics, while a large one would induce noise. 
We can see from Figure \ref{fig:title_freq} that the frequency of section labels is long-tailed, thus retaining only a small portion is able to capture the major part of information.
Table \ref{tab:section_no_comparison} records the experiment results we conducted on domain \textit{Company}. $n_t = 20$ reaches a peak on ROUGE 1, 2 and L scores, indicating that 20 is a reasonable number of section labels.
\begin{table}[htbp]
    \centering
    \begin{tabular}{c|c c c}
        \toprule
       \#Labels & R1 & R2 & RL \\
        \midrule 
        10 & .337 & .117 & .312 \\
        20 & \textbf{.340} & \textbf{.118} & \textbf{.315} \\
        30 & .336 & .117 & .311 \\
        \bottomrule
    \end{tabular}
    \caption{ROUGE F1 scores of different $n_t$.}
    \label{tab:section_no_comparison}
\end{table}

\subsection{Case Study}
Table \ref{tab:qualitative} shows the generated Wikipedia abstracts by different models about film \textit{Majina There}.
We can see that the gold abstract contains information about three topics: basic information (region, director, and producer), actors, and music.

Among the models, TF-S2S produces an abstract with a proper pattern but contains wrong information and BART misses the musical information topic.
CV-S2D+T, HierSumm, and our TWAG model both cover all three topics in the gold abstract, however, CV-S2D+T makes several factual errors like the release date and actors and HierSumm suffers from redundancy.
TWAG covers all three topics in the gold abstract and discovers extra facts, proving itself to be competent in generating comprehensive abstracts.

\begin{table*}[htbp]
    \centering
    \begin{tabular}{c|c c|c c|c c}
        \toprule
        \multirow{2}{*}{Model} & \multicolumn{2}{c|}{Company} & \multicolumn{2}{c|}{Film} & \multicolumn{2}{c}{Animal} \\
        & Score & Non-0 & Score & Non-0 & Score & Non-0 \\
        \midrule 
        TF-S2S & .075 & .694  & .000 & .000 & .000 & .000 \\
        CV-S2D+T & .237 & .660 & .040 & .143 & .382 & .576 \\
        HierSumm & .255 & .896 & .213 & .327 & .000 & .000 \\
        BART & .591 & .813 & .452 & .796 & .342 & .653 \\
        TWAG (ours) & \textbf{.665} & \textbf{.903} & \textbf{.669} & \textbf{.918} & \textbf{.543} & \textbf{.868} \\
        \bottomrule
    \end{tabular}
    \caption{Human evaluation results in QA scheme. Score represents the mean score and non-0 represents the percentage of answered questions.}
    \label{tab:human_qa_results}
\end{table*}

\begin{table*}[htbp]
    \centering
    \begin{tabular}{c|c c c|c c c|c c c}
        \toprule
        \multirow{2}{*}{Model} & \multicolumn{3}{c|}{Company} & \multicolumn{3}{c|}{Film} & \multicolumn{3}{c}{Animal} \\
        & C & F & S & C & F & S & C & F & S \\
        \midrule 
        TF-S2S & 2.69 & 2.71 & 2.67 &1.93 & 2.71 & 2.84 & 2.22 & 2.96 & 2.76 \\
        CV-S2D+T & 2.42 & 2.36 & 2.73 & 2.29 & 2.69 & 2.98 & 2.80 & 3.18 & 3.18 \\
        HierSumm & \textbf{2.96} & 2.64 & 1.69 & 3.13 & 2.78 & 2.04 & 2.80 & 3.13 & 1.82 \\
        BART & 2.64 & 2.82 & \textbf{ 3.00} & 2.87 & 3.02 & 3.24 & 2.78 & 3.11 & 3.00 \\
        TWAG (ours) & 2.91 & \textbf{2.87} &2.91 & \textbf{3.20} & \textbf{3.16} & \textbf{3.44} & \textbf{3.56} & \textbf{3.58} & \textbf{3.40} \\
        \bottomrule
    \end{tabular}
    \caption{Human evaluation results in linguistic quality scoring. C indicates completeness, F indicates fluency and S indicates succinctness.}
    \label{tab:human_linguistic_results}
\end{table*}

\subsection{Human Evaluation}
We follow the experimental setup of \cite{perez2019generating} and conduct a human evaluation consisting of two parts.
A total of 45 examples (15 from each domain) are randomly selected from the test set for evaluation.

The first part is a question-answering (QA) scheme proposed in \cite{clarke2010discourse} in order to examine factoid information in summaries.
We create 2-5 questions\footnote{Example questions are listed in the Appendix C, and the whole evaluation set is included in the our code repository.} based on the golden summary which covers the appeared topics, and invite 3 participants to answer the questions by taking automatically-generated summaries as background information.
The more questions a summary can answer, the better it is. To quantify the results, we assign a score of 1/0.5/0.1/0 to a correct answer, a partially correct answer, a wrong answer and those cannot be answered, and report the average score over all questions.
Notice that we give a score of 0.1 even if the participants answer the question incorrectly, because a wrong answer indicates the summary covers a certain topic and is superior to missing information.
Results in Table \ref{tab:human_qa_results} shows that 1) taking summaries generated by TWAG is capable of answering more questions and giving the correct answer, 2) TF-S2S and HierSumm perform poorly in domain \textit{Film} and \textit{Animal}, which is possibly a consequence of under-fitting in small datasets.

The second part is an evaluation over linguistic quality.
We ask the participants to read different generated summaries from 3 perspectives and give a score of 1-5 (larger scores indicates higher quality):
\textbf{Completeness} (does the summary contain sufficient information?), \textbf{Fluency} (is the summary fluent and grammatical?) and \textbf{Succinctness} (does the summary avoid redundant sentences?)
Specifically, 3 participants are assigned to evaluate each model, and the average scores are taken as the final results.
Table \ref{tab:human_linguistic_results} presents the comparison results, from which we can see that, the linguistic quality of TWAG model outperforms other baseline models, validating its effectiveness.

\section{Conclusion}
\label{sec:conclusion}
In this paper, we propose a novel topic-guided abstractive summarization model \modelnames for generating Wikipedia abstracts.
It investigates the section labels of Wikipedia, dividing the input document into different topics to improve the quality of generated abstract.
This approach simulates the way how human recognize entities, and experimental results show that our model obviously outperforms existing state-of-the-art models which view Wikipedia abstracts as plain text.
Our model also demonstrates its high data efficiency.
In the future, we will try to incorporate pretrained language models into the topic-aware abstract generator module, and apply the topic-aware model to other texts rich in topical information like sports match reports.

\section*{Acknowledgments}

We thank the anonymous reviewers for their insightful comments. This work is supported by the National Key Research and Development Program of China (2017YFB1002101), NSFC Key Project (U1736204) and a grant from Huawei Inc. 

\clearpage

\section*{Ethical Considerations}
\modelnames could be applied to applications like automatically writing new Wikipedia abstracts or other texts rich in topical information.
It can also help human writers to examine whether they have missed information about certain important topics.

The benefits of using our model include saving human writers' labor and making abstracts more comprehensive.
There are also important considerations when using our model.
Input texts may violate copyrights when inadequately collected, and misleading texts may lead to factual mistakes in generated abstracts.
To mitigate the risks, researches on how to avoid copyright issues when collecting documents from the Internet would help.

\bibliographystyle{acl_natbib}
\bibliography{acl2021}

\appendix

\section{Topic Allocation}
For each domain, we sort section labels by frequency and choose the top $n_t = 20$ frequent section labels, then manually allocate them into different topics.
Section labels with little semantic information like \textit{Reference} and \textit{Notes} are discarded in allocation to reduce noise.
Table \ref{tab:topic_detail} shows how we allocate section labels into topics in domain \textit{Company}, \textit{Film} and \textit{Animal}.

An additional \textit{NOISE} topic is added to each domain to detect website noises.
We build training records for \textit{NOISE} by finding noise text in the training set of WikiCatSum by regular expressions.
For example, we view all text containing ``cookie'', ``href'' or text that seems to be a reference as noise.

\section{Trivia about Baselines}
We use BART-base as the baseline for comparison because BART-large performs poorly in experiments.
BART-large starts generating redundant results when using only 4\% training data, and its training loss also decreases much slower than BART-base.
We infer that BART-large may overfit on training data, and BART-base is more competent to be the baseline.

\section{Human Evaluation Example}
Table \ref{tab:he_example} shows an example of gold summary, its corresponding question set and system outputs.
The full dataset we used for human evaluation can be found in our code repository.

\begin{table*}[htbp]
    \centering
    \begin{tabular}{|c|c|c|}
        \hline
        \textbf{Domain} & \textbf{Topic} &\textbf{ Section Labels} \\
        \hline 
        \multirow{6}{*}{Company} & History & \textcolor{Orange}{History}, \textcolor{Green}{Company history}, \textcolor{Blue}{Ownership} \\ \cline{2-3}
                                 & \multirow{2}{*}{Product} & \textcolor{Orange}{Products}, \textcolor{Orange}{Services}, \textcolor{Orange}{Destinations}, \\
                                 & & \textcolor{Green}{Products and services}, \textcolor{Green}{Technology} \\ \cline{2-3}
                                 & Location & \textcolor{Orange}{Fleet}, \textcolor{Orange}{Operations}, \textcolor{Green}{Subsidiaries} , \textcolor{Green}{Locations} \\ \cline{2-3}
                                 & \multirow{2}{*}{Reception} & \textcolor{Orange}{Awards}, \textcolor{Blue}{Controversies}, \textcolor{Blue}{Controversy}, \\
                                 & & \textcolor{Blue}{Criticism}, \textcolor{Blue}{Accidents and incidents}, \textcolor{Blue}{Reception} \\ \cline{2-3}
        \hline
        \multirow{6}{*}{Film}    & Cast & Cast, Casting \\ \cline{2-3}
                                 & Plot & Plot, Synopsis, Plot summary \\ \cline{2-3}
                                 & Production & Production, Filming, Development \\ \cline{2-3}
                                 & \multirow{2}{*}{Reception} & Reception, Critical reception, Critical response,  \\ 
                                 &  & Awards, Accolades, Awards and nominations \\ \cline{2-3}
                                 & Box office & Box office \\ \cline{2-3}
        \hline
        \multirow{5}{*}{Animal}  & Distribution & Distribution \\ \cline{2-3}
                                 & Taxonomy & Taxonomy, Species, Subspecies, Classification \\ \cline{2-3}
                                 & \multirow{2}{*}{Description} & Description, Habitat, Ecology, Behaviour, Biology, \\
                                 & & Diet, Feeding, Breeding, Reproduction, Life cycle \\ \cline{2-3}
                                 & Conservation Status & Status, Conservation, Conservation status \\ \cline{2-3}
        \hline
    \end{tabular}
    \caption{Detailed allocation of section labels. In domain \textit{Company}, \textcolor{Orange}{Orange} labels are labels selected when $n_t = 10$, \textcolor{Green}{Green} labels are additional labels selected when $n_t = 20$, and \textcolor{Blue}{Blue} labels are additional labels selected when $n_t = 30$.}
    \label{tab:topic_detail}
\end{table*}

\begin{table*}[htbp]
  \small
  \centering
    \begin{tabular}{|p{0.9\linewidth}|}
    \hline
    {\textbf{\textit{Gold Abstract:}} l'avare is a 1980 french comedy film written and directed by louis de funès and jean girault , and starring de funès. the english title of the film is the miser. it is an adaptation of molière 's famous comedy l'avare ou l'école du mensonge ( the miser ). de funès tried to draw out the unhappy side of the character. harpagon , unloved by humanity, is driven to an obsessive love of money. } \\
    \hline
    \makecell[l]{\textbf{\textit{Questions:}} \\ 1. When and where was l'avare released? \\ 2. Who stars in l'avare? \\ 3. Who directed l'avare? \\ 4. What is the English name of l'avare? } \\
    \hline
    {\textbf{\textit{TF-S2S:}} der er et yndigt land is a 1983 danish drama film directed by morten arnfred. it was entered into the 33rd berlin international film festival , where it won an honourable mention.} \\
    \hline
    {\textbf{\textit{CV-S2D+T:}} \&lt;unk\&gt; 's \&lt;unk\&gt; is a french comedy film from 1954, directed by jean girault, starring jean marais and louis de funès. it was screened in the un certain regard section at the 2015 cannes film festival.} \\
    \hline
    {\textbf{\textit{HierSumm:}}( hangul ; rr : l'am ) is a french drama film directed by louis de funès. it is based on a play by molière. it stars louis de funès. it was entered into the 36th berlin international film festival. the film was nominated for the golden globe award for best foreign language film. it was also nominated for the golden globe award for best foreign language film. \ldots} \\
    \hline
    {\textbf{\textit{BART:}} l'avare ( english : the miser ) is a 1980 french drama film directed by louis de funès and starring jean girault. it was based on an original screenplay co-written with julien françois .} \\
    \hline
    {\textbf{\textit{TWAG:}} the miser ( french : l’avare ) is a 1980 french drama film directed by funes de funès. the film stars louis de funès , sanctioning cléante , and broach harpagon.} \\
    \hline
    \end{tabular}
  \caption{Example of Gold summary, question set and system outputs for the QA evaluation study.}
  \label{tab:he_example}
\end{table*}

\end{document}